\pdfoutput=1

\documentclass[11pt]{article}

\usepackage[]{acl}

\usepackage{times}
\usepackage{latexsym}

\usepackage[T1]{fontenc}

\usepackage[utf8]{inputenc}

\usepackage{microtype}

\usepackage{graphicx}
\usepackage{amsmath}
\usepackage{enumitem}
\usepackage{booktabs}
\usepackage{multirow}
\usepackage{amssymb}
\usepackage{xspace}

\newcommand{\ft}{\textit{Fine-tuning}\xspace}
\newcommand{\ewc}{\textit{EWC}\xspace}
\newcommand{\replay}{\textit{Replay}\xspace}
\newcommand{\adapter}{\textit{AdapterCL}\xspace}
\newcommand{\pt}{\textit{Prompt Tuning}\xspace}
\newcommand{\cpt}{\textit{Continual Prompt Tuning}\xspace}

\newcommand{\multipt}{\textit{Multi-task Prompt Tuning}\xspace}
\newcommand{\wmem}{\textit{w/ memory}\xspace}
\newcommand{\wmemback}{\textit{w/ memory \& backward}\xspace}

\newcommand{\clinit}{CLInit\xspace}
\newcommand{\selectinit}{SelectInit\xspace}

\title{Continual Prompt Tuning for Dialog State Tracking}

\author{Qi Zhu$^1$, Bing Li$^1$, Fei Mi$^2$, Xiaoyan Zhu$^1$, Minlie Huang$^1$\footnotemark[1] \\
$^1$CoAI Group, DCST, IAI, BNRIST, Tsinghua University\\
$^2$Huawei Noah’s Ark Lab \\
{\small \tt zhu-q18@mails.tsinghua.edu.cn, aihuang@tsinghua.edu.cn} \\}

\begin{document}
\maketitle

\renewcommand{\thefootnote}{\fnsymbol{footnote}}
\footnotetext[1]{Corresponding author.}
\renewcommand{\thefootnote}{\arabic{footnote}}

\begin{abstract}
A desirable dialog system should be able to continually learn new skills without forgetting old ones, and thereby adapt to new domains or tasks in its life cycle. However, continually training a model often leads to a well-known catastrophic forgetting issue. In this paper, we present Continual Prompt Tuning, a parameter-efficient framework that not only avoids forgetting but also enables knowledge transfer between tasks. To avoid forgetting, we only learn and store a few prompt tokens' embeddings for each task while freezing the backbone pre-trained model. To achieve bi-directional knowledge transfer among tasks, we propose several techniques (continual prompt initialization, query fusion, and memory replay) to transfer knowledge from preceding tasks and a memory-guided technique to transfer knowledge from subsequent tasks. Extensive experiments demonstrate the effectiveness and efficiency of our proposed method on continual learning for dialog state tracking, compared with state-of-the-art baselines.
\end{abstract}

\section{Introduction}
Recently, most studies have focused on developing dialog systems for specific domains in an offline manner, assuming the data distribution stays the same.
However, this is far from realistic because a deployed dialog system is often required to support new domains and provide more services constantly over time.
Therefore, it is crucial for a dialog system to continually learn new tasks without forgetting old ones with high efficiency. 

Previous studies on continual learning \cite{kirkpatrick2017ewc,li2017lwf} mainly focused on solving the \textit{catastrophic forgetting (CF)} problem \cite{mccloskey1989catastrophic}: 
when a neural model is trained on a sequence of tasks, new tasks may interfere catastrophically with old tasks.
Simply storing a model version for each task to mitigate forgetting is prohibitive as the number of tasks grows, especially when the model size is large.
To mitigate catastrophic forgetting with low computation and storage overhead, recent methods freeze the backbone model and propose to train a weight/feature mask \cite{mallya2018piggyback,geng-etal-2021-continual} or an adapter \cite{madotto-etal-2021-continual} for each task independently.
However, the techniques above are still not efficient enough, and they largely ignore knowledge transfer among tasks.


\begin{figure}[t]
    \centering
    \includegraphics[width=\linewidth]{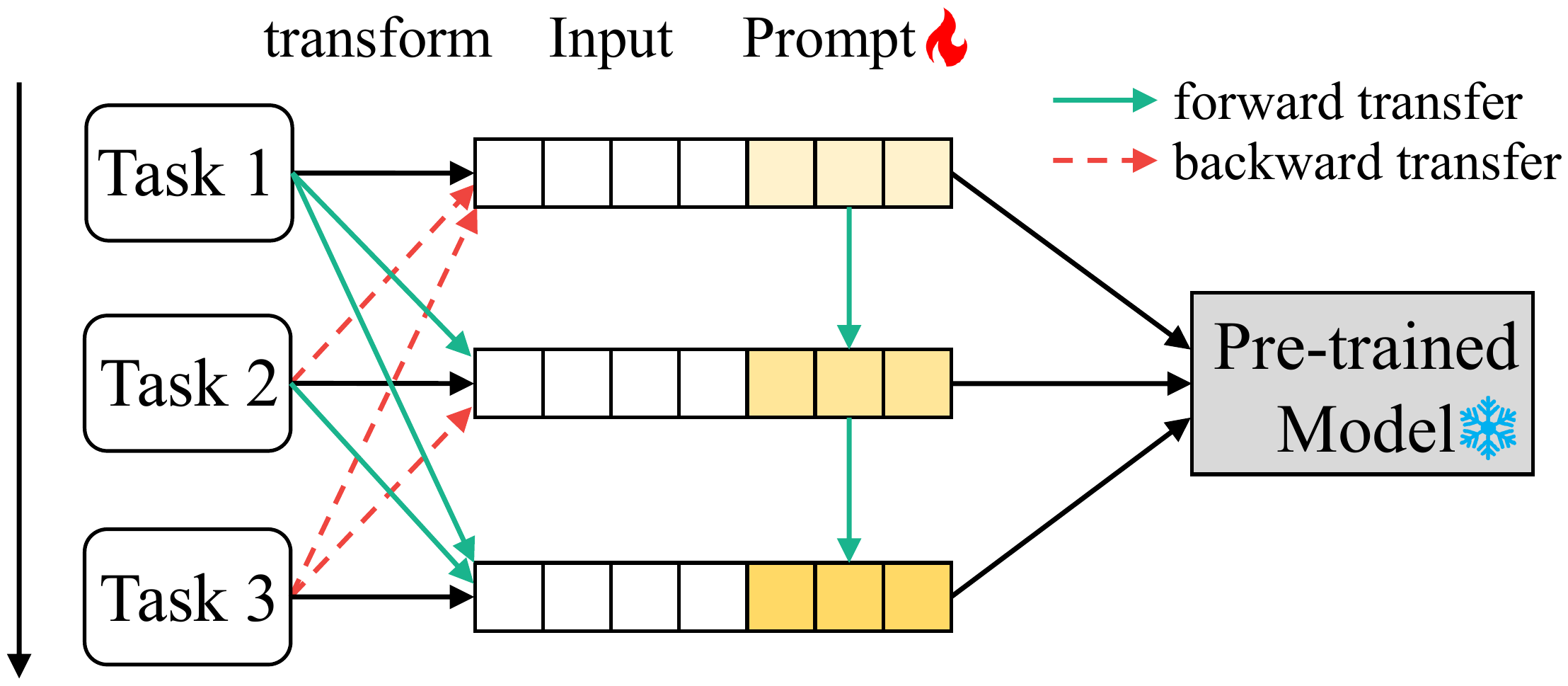}
    \caption{An illustration of \cpt. We train a soft prompt for each task and freeze the pre-trained model. 
    Several techniques are proposed to transfer knowledge from preceding tasks (green solid arrows) and subsequent tasks (red dashed arrows).}
    \vspace{-1em}
    \label{fig:1}
\end{figure}

In this paper, we develop prompt tuning \cite{lester-etal-2021-power} for continual learning.
We freeze the backbone pre-trained model and train a few prompt tokens' embeddings for each task, which is highly parameter-efficient to avoid forgetting.
As illustrated by yellow components in Figure \ref{fig:1}, we concatenate the input with a few \textit{tunable} task-specific prompt tokens before feeding it to a \textit{frozen} pre-trained model. 
Since these prompt tokens have only a small number of parameters (0.1\% of the pre-trained model's parameters in our experiments), we can efficiently train and store the prompt for each task.
During inference, the same pre-trained model can handle different tasks by inputting different prompts, which is friendly for deployment.

Unlike the vanilla approach of training each task's prompt from scratch and fixing it afterward, we propose \cpt, a framework that enables \textbf{knowledge transfer} between tasks. 
We consider transferring knowledge from both preceding tasks (forward) and subsequent tasks (backward). 
To realize forward transfer, we propose several techniques, including continual prompt initialization, query fusion, and memory replay (green solid arrows in Figure \ref{fig:1}). 
To achieve positive backward transfer, we propose a memory-guided technique that uses subsequent tasks' data to update the previous tasks' prompts selectively (red dashed arrows in Figure \ref{fig:1}).

We conduct experiments on Dialog State Tracking (DST), a core component of a dialog system, using the Schema-Guided Dialog dataset \cite{rastogi2020scalable}.
The model continually learns new services that have multiple slots to fill.
We concatenate all slots' descriptions with the input and insert a sentinel token after each description, formulating DST as a masked spans recovering task, which is similar to the pre-training objective of T5 \cite{raffel2020t5}.
We empirically show that our proposed framework effectively outperforms state-of-the-art baselines on continual learning for DST, and is extremely efficient in terms of computation and storage.\footnote{Code and data are publicly available at \url{https://github.com/thu-coai/CPT4DST}}

To summarize, our main contributions are:
\begin{enumerate}[itemsep=0pt,topsep=2pt,leftmargin=12pt]
    \item For the first time, we develop prompt tuning for continual learning, which avoids forgetting efficiently and is friendly for deployment.
    
    \item We investigate several techniques for forward and backward knowledge transfer based on prompt tuning, further boosting the continual learning performance.
    
    \item Our experiments on continual DST demonstrate the superior performance and efficiency of our proposed method.
\end{enumerate}

\section{Related Work}
\subsection{Continual Learning}
Continual Learning (CL) studies the problem of continually acquiring knowledge from a data stream and reusing it for future learning while avoiding forgetting.
Three kinds of CL methods have been developed.
\textit{Rehearsal} methods store and replay some training samples from previous tasks
\cite{rebuffi2017icarl,lopez2017gem}. 
\textit{Regularization} methods apply additional loss to aid knowledge consolidation
\cite{kirkpatrick2017ewc,li2017lwf}. 
\textit{Architectural} methods introduce task-specific parameters for new tasks and fix parameters for old tasks to prevent forgetting, to which our method belongs.
Previous architectural methods include dynamic expanding network structure \cite{rusu2016progressive}, iterative network pruning and re-training \cite{mallya2018packnet}, learning a parameter mask for each task individually \cite{mallya2018piggyback}, etc.

For continual learning in dialog system, variants of general CL methods have been applied
\cite{lee2017toward,shen-etal-2019-progressive,wu-etal-2019-transferable,mi-etal-2020-continual,geng-etal-2021-continual}.
AdapterCL \cite{madotto-etal-2021-continual} is the most related to our work, which freezes the pre-trained model and learns an adapter \cite{houlsby2019adapter} for each task independently.
Compared with AdapterCL, our method is more parameter-efficient, and we explore the effect of both forward and backward transfer.

\subsection{Prompt-based Tuning}
Recent studies have found that using a textual prompt to convert downstream tasks to the language modeling task is a more effective way to use pre-trained language models than typical fine-tuning \cite{brown2020gpt3,schick-schutze-2021-just}.
Prompts can be manual designed \cite{petroni-etal-2019-language} or generated automatically \cite{shin-etal-2020-autoprompt,jiang-etal-2020-know,gao-etal-2021-making}. 
Since searching prompts in discrete spaces is sub-optimal, some works \cite{qin-eisner-2021-learning,liu2021gpt,han2021ptr} combine hard text prompts and soft prompts whose embeddings are learned through back-propagation.
\citet{lester-etal-2021-power} show that freezing the pre-trained model and only tuning soft prompts, known as prompt tuning, is parameter-efficient and becomes more competitive with fine-tuning as the model size grows.

Prompt tuning differs from embedding adapter \cite{zhu-etal-2021-counter-interference} that aims to address the multilingual embedding deficiency.
An embedding adapter transforms all tokens embeddings but do not affect transformer layers' computation, while prompt tuning does not change tokens embeddings but adds new tunable prompt tokens to the input, serving as context and affecting all following transformer layers.
\citet{gu2021ppt} and \citet{vu2021spot} further explore the transferability of soft prompts across tasks.
While they investigate one-step adaptation, we are interested in prompt transfer in the continual learning setting.

\subsection{Dialog State Tracking}
Dialog State Tracking (DST) aims to capture user goals in the form of (slot, value) pairs. 
Traditional ontology-based classification methods \cite{mrksic-etal-2017-neural,lee-etal-2019-sumbt} require access to all candidate values. 
To alleviate the reliance on the ontology and improve generalization to unseen values, some work extract values from a dialog context \cite{xu-hu-2018-end,gao-etal-2019-dialog} while
others generate values directly to handle situations where values are missing from the context \cite{wu-etal-2019-transferable,hosseiniasl2020simple}.

Generation-based models either generate all (slot, value) pairs in one pass \cite{hosseiniasl2020simple,madotto-etal-2021-continual} or generate value for each given slot separately \cite{wu-etal-2019-transferable}.
The former are more efficient but can only predict in-domain slots and lack transferability while the latter can incorporate more information about a slot as a query, such as a brief natural language description \cite{rastogi2020scalable}, slot type information \cite{lin-etal-2021-leveraging}, possible values \cite{lee-etal-2021-dialogue}, and the task definition and constraint \cite{mi2021cins}.
Our proposed method integrates multiple slot descriptions into a single query and generates all values in one pass, which improves performance without losing efficiency.

\section{Method}

\begin{figure*}[t]
    \centering
    \includegraphics[width=\linewidth]{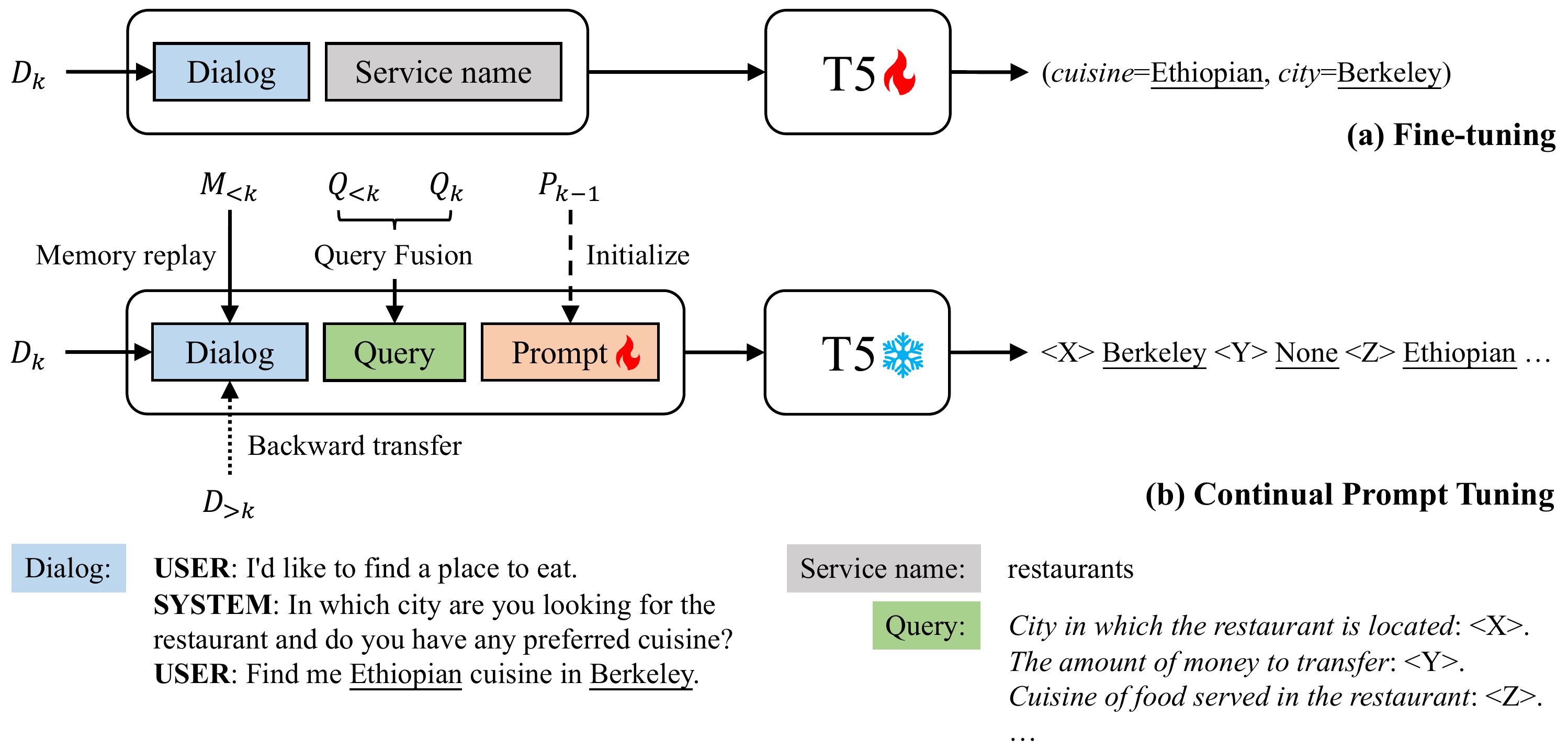}
    \vspace{-2em}
    \caption{
    An illustration of \ft and \cpt for continual DST. 
    \textbf{(a) \ft} takes the dialog and current service's name as input and tunes T5 to generate slot-value pairs.
    \textbf{(b) \cpt} feeds the dialog, query consisting of slot descriptions and sentinel tokens, and prompt tokens to frozen T5 and tunes the prompt's embeddings to generate values for all slots in the query.
    Continual prompt initialization, query fusion, and memory replay are proposed to enhance forward transfer while subsequent services' data will be used for backward transfer.
    We show an example dialog, service name, fused query, and expected outputs. Slot \textit{names} and \textit{descriptions} are in italic and \underline{values} are underlined. 
    Note that the second slot description in the query belongs to another service ("banks") and is inserted by query fusion.
    }
    \label{fig:2}
    \vspace{-1em}
\end{figure*}

\subsection{Overview}
The goal of continual learning is to sequentially learn a model $f:\mathcal{X} \times \mathcal{T} \to \mathcal{Y}$ from a stream of tasks $\mathcal{T}_1...\mathcal{T}_T$ that can predict the target $y$ given the input $x$ and task $\mathcal{T}_k \in \mathcal{T}$.
We denote the data for each task $\mathcal{T}_k$ as $D_k$.
Our method is based on pre-trained language models.
Instead of fine-tuning a pre-trained model in a traditional manner (Figure \ref{fig:2}(a)), we freeze the model but "reprogram" it to solve task $\mathcal{T}_k$ by adding $m$ new soft prompt tokens $P_k=P_k^1P_k^2...P_k^m$ to the textual input and tuning the embeddings of $P_k$ only.
Since the prompt's parameters are much less than the model's, we save $P_k$ for each task to avoid forgetting.

We treat each service/API as a task in continual DST (service and task are used interchangeably).
To incorporate informative slot descriptions and ease the decoding process, we convert the descriptions into a query with masked spans and formulate DST as a masked spans recovering task (Sec. \ref{sec:format}).
To enhance knowledge transfer between tasks, we propose continual prompt initialization, query fusion, and memory replay for forward transfer (Sec. \ref{sec:fwt}) and explore a memory-guided technique for backward transfer (Sec. \ref{sec:bwt}).

\subsection{DST as Masked Spans Recovering}
\label{sec:format}
In DST, each service $\mathcal{T}_k$ has a set of pre-defined slots $\mathcal{S}_k=\{s_1,...,s_{n_k}\}$ to be tracked.
The input $x$ is a dialog and the output $y$ consists of slot-value pairs: $\{(s_1,v_1),(s_2,v_2),...,(s_{n_k},v_{n_k})\}$.
Similar to many NLP tasks, DST can be formulated as a text-to-text generation task.
Formally, we define a function $g_k:\mathcal{X} \times \mathcal{Y} \to \mathcal{V^*} \times \mathcal{V^*}$ for each service $\mathcal{T}_k$ to transform the original data $(x,y)$ to:
\begin{equation}
    \tilde{x}, \tilde{y} = g_k(x, y)
\end{equation}
where $\mathcal{V}$ is the vocabulary and $\tilde{x}, \tilde{y}$ are texts that serve as the model input and output, respectively. 
For example, $\tilde{x}$ can be the concatenation of $x$ and service name, while $\tilde{y}$ is a sequence of slot-value pairs \cite{madotto-etal-2021-continual} (Figure \ref{fig:2}(a)).

Previous research has shown that incorporating a natural language description $d_i$ for each slot $s_i$ is beneficial \cite{lin-etal-2021-leveraging,lee-etal-2021-dialogue}.
They concatenate the dialog $x$ with each slot description $d_i$ and decode the value $v_i$ independently. 
However, separately decoding is inefficient, especially when there are many slots.
To solve this, we concatenate all slot descriptions and insert a sentinel token after each description to form a query added to the input, formulating DST as a masked spans recovering task that generates all slot values in one pass:
\begin{equation}
\begin{split}
    \tilde{x} &= [x; Q_k; P_k]\\
    Q_k &= ``d_1^k : \langle \mathrm{M}_1 \rangle . \ ... \ d_{n_k}^k : \langle \mathrm{M}_{n_k} \rangle ."\\
    \tilde{y} &= ``\langle \mathrm{M}_1 \rangle \ v_1^k \ ... \langle \mathrm{M}_{n_k} \rangle \ v_{n_k}^k"
\end{split}
\end{equation}
where $[\cdot ; \cdot]$ is the concatenation operation and $\langle \mathrm{M}_* \rangle$ are distinct sentinel tokens representing masked spans.
The \textbf{query} $Q_k$ contains all $n_k$ slot descriptions for task $\mathcal{T}_k$ with $n_k$ masked spans and $\tilde{y}$ contains corresponding slot values leaded by the sentinel tokens.
If the value of a slot can not be inferred from the input, we set it to "None".
We freeze the pre-trained model's parameters $\theta$ and only optimize the prompt's parameters $\theta_{P_k}$ for each service $\mathcal{T}_k$.
The loss function is:
\begin{equation}
    \mathcal{L}_{\theta_{P_k}}(D_k) = - \sum_{j=1}^{|D_k|} \log p_{\theta}(\tilde{y}_j^k|[x_j^k; Q_k; P_k])
\end{equation}

\subsection{Forward Transfer}
\label{sec:fwt}
Reusing the knowledge acquired from preceding tasks often improves and accelerates the learning on future tasks.
Therefore, we propose three types of techniques for forward transfer that can be employed in combination.

\subsubsection{Continual Prompt Initialization}
\label{sec:init}
An intuitive way to transfer knowledge is parameter initialization.
We explore two continual prompt initialization strategies. \textbf{\clinit} uses last task's prompt $P_{k-1}$ to initialize current task's prompt $P_k$.
\textbf{\selectinit} evaluates all $\{P_{j}\}_{j<k}$ on the validation set of $\mathcal{T}_k$ without training and selects the one with the lowest loss to initialize $P_k$.
The initial prompt of \clinit has been continually trained on all previous tasks, while \selectinit only considers the most relevant task without interference from its subsequent tasks.
We empirically compare these two strategies in Sec. \ref{sec:exp_init}.

\subsubsection{Query Fusion}
\label{sec:aug}
We hope the model can learn to generate values according to any slot descriptions, which is a general skill that may improve performance on future tasks.
However, when training on the current task, there is only one query that consists of the slot descriptions of that task in a fixed order, which may hinder the model from learning the general skill.
Therefore, we propose to augment the query by mixing slot descriptions from the current and previous tasks to help the prompt better understand the correspondence between slot descriptions and values.
We fuse the query $Q_k$ with previous tasks' queries $\{Q_j\}_{j<k}$ for each sample, including three steps: 1) sample $n_1$ slots from $\mathcal{S}_k$ randomly, where $n_1$ is sampled from $[1, |\mathcal{S}_k|]$ uniformly. 
2) sample $n_2$ slots from previous tasks' slots $\bigcup_{i<k} \mathcal{S}_i$ randomly, where $n_2$ is sampled from $[1, n_1]$ uniformly. 
3) combine the above $n_1$ and $n_2$ slots' descriptions in a random order as new $Q_k^{'}$, and modify $\tilde{y}$ accordingly.
Note that some original slots are dropped, and values for added slots are set to "None".

\subsubsection{Memory Replay}
\label{sec:replay}
Previous studies \cite{rebuffi2017icarl,lopez2017gem} store a few samples for each task and replay them when training on new tasks to mitigate forgetting.
Since our prompt tuning framework has already resolved forgetting, we focus on how these samples benefit the current task.
We assume we can store $|M|$ samples for each task ($|M|$ should be small) and denote $M_i$ as the memory for task $\mathcal{T}_i$.
When a new task $\mathcal{T}_k$ comes, we optimize $P_k$ on $D_k$ and $M_{<k}=\bigcup_{i<k}M_i$ jointly, changing the loss function to $\mathcal{L}_{\theta_{P_k}}(D_k+M_{<k})$.

When combined with query fusion, query $Q_i$ for samples in the memory $M_i$ are also fused with queries $\{Q_j\}_{j\leq k,j\neq i}$ from other seen tasks, including the current task.
Note that in this way, samples from other tasks can be viewed as "positive" samples to those added slots in $Q_i^{'}$ since these samples may have not "None" values for those added slots.

\subsection{Memory-Guided Backward Transfer}
\label{sec:bwt}
Although fixing $P_k$ immediately after training on task $\mathcal{T}_k$ can avoid forgetting, it also blocks the backward knowledge transfer from future tasks.
Motivated by \citet{chaudhry2018agem}, we explore whether it is possible to improve the performance on previous tasks with the help of memory when a new task comes.
Specifically, for each previous task $\mathcal{T}_i, i<k$, we initialize a new prompt $P_i^{(k)}$ to $P_i$ and trained it on current task's data $D_k$ with memory $M_i$ as regularization.
During training, we sample a batch from $D_k$ and a batch from $M_i$ synchronously and denote the gradient from each batch as $g_{ori}$ and $g_{ref}$, respectively.
We decide the gradient for update according to the angle between $g_{ori}$ and $g_{ref}$:
\vspace{-0.5em}
\begin{equation}
    g = \begin{cases}
    g_{ori}, &\text{if } g_{ori}^T\ g_{ref} > 0 \\
    0, &\text{otherwise}
    \end{cases}
\vspace{-0.5em}
\end{equation}
which means we abort the update that will increase the loss on memory batch.
We empirically find that this simple abortion is better than projecting $g_{ori}$ onto the normal plane of $g_{ref}$ \cite{chaudhry2018agem}.
After training, we update $P_i$ to $P_i^{(k)}$ if $P_i^{(k)}$ obtains lower loss and better (or equal) performance on $M_i$ than $P_i$.

\section{Experimental Setup}
Recently, \citet{madotto-etal-2021-continual} proposed a continual learning benchmark for task-oriented dialog systems and compared several classic CL methods.
We adapt their data processing steps and baselines in our experiments.

\subsection{Dataset}
We conduct experiments on Schema-Guided Dialog dataset (SGD) \cite{rastogi2020scalable} that has 44 services over 19 domains. It also provides a one-sentence description for each slot.
We treat each service as a task and only consider dialogs involving a single service.
We randomly split a service's dialogs into train/val/test sets at the ratio of 7:1:2.
The number of training samples of each service ranges from 112 to 4.7K, and there are 2 to 10 slots for one service.
More details about data statistics can be found in the Appendix (Table \ref{tab:dataset_count}).

\subsection{Evaluation Protocol}
We evaluate DST performance using the widely adopted Joint Goal Accuracy (JGA) \cite{wu-etal-2019-transferable}, which requires all slots' values are correctly predicted.
We assign the target service during testing to avoid ambiguity since the same dialog can be parsed differently under different services.
We denote $a_{j,i}$ as the JGA on the test set of task $\mathcal{T}_i$ right after training on task $\mathcal{T}_j$.
We evaluate the CL performance as the average JGA on all tasks after training on the final task $\mathcal{T}_T$: 
\vspace{-0.75em}
\begin{equation}
    \mathbf{Avg. \ JGA}=\frac{1}{T}\sum_{i=1}^Ta_{T,i}
\vspace{-0.75em}
\end{equation}

Following \citet{lopez2017gem}, we define two metrics to measure the effect of forward transfer and backward transfer, respectively:
\vspace{-0.75em}
\begin{equation}
\begin{aligned}
    \mathbf{FWT} &=\frac{1}{T-1}\sum_{i=2}^{T}a_{i-1,i} \\
    \mathbf{BWT} &=\frac{1}{T-1}\sum_{i=1}^{T-1}a_{T,i}-a_{i,i}
\end{aligned}
\vspace{-0.75em}
\end{equation}
FWT is the averaged zero-shot performance on new tasks, evaluating a model's generalization ability.
BWT assesses the impact that learning on subsequent tasks has on a previous task.
Negative BWT indicates that the model has forgotten some previously acquired knowledge.

\subsection{Baselines and Training Details}
We adopt the following models from \citet{madotto-etal-2021-continual} as baselines:
\begin{itemize}[leftmargin=*,itemsep=2pt,topsep=0pt,parsep=0pt]
    \item \textbf{\ft:} Fine-tune the model on new task data continually.
    \item \textbf{\replay:} Save $|M|$ samples randomly sampled from the training set of each task $\mathcal{T}_i$ to memory $M_i$ and jointly train the model on new task data $D_k$ and memory $M_{<k}$.
    \item \textbf{\ewc:} Maintain the memory in the same way as \replay but use it to compute the Fisher information matrix for regularization \cite{kirkpatrick2017ewc}.
    \item \textbf{\adapter:} Freeze the pre-trained model and train a residual Adapter \cite{houlsby2019adapter} for each task independently \cite{madotto-etal-2021-continual}.
\end{itemize}
Above methods use the same input and output format as in Figure \ref{fig:2}(a).

Prompt tuning based methods including our proposed \cpt are list below:
\begin{itemize}[leftmargin=*,itemsep=2pt,topsep=0pt,parsep=0pt]
    \item \textbf{\pt:} Formulate DST as a masked spans recovering task (Sec. \ref{sec:format}) and only tune the prompt for each task independently.
    \item \textbf{\multipt:} \pt in a multi-task manner instead of CL. Train a single prompt using all tasks' data concurrently.
    \item \textbf{\cpt:} \pt with \clinit (Sec. \ref{sec:init}) and query fusion (Sec. \ref{sec:aug}).
    \begin{itemize}[leftmargin=*,itemsep=0pt,topsep=0pt,parsep=0pt]
        \item \textbf{\wmem} with memory replay (Sec. \ref{sec:replay}).
        \item \textbf{\wmemback} with memory replay and memory-guided backward transfer (Sec. \ref{sec:bwt}).
    \end{itemize}
\end{itemize}

We use the following setting in the experiments unless otherwise specified.

\paragraph{Training task sequences}
Since a sequence of all (44) tasks is too long for the evaluation purpose, we conduct most of the experiments on 15 tasks chosen at random to save computing resources.
We run \adapter, \pt, and \multipt 5 times with different random seeds because they are agnostic to task order.
The FWT and BWT metrics for these models are left blank.
We run other methods in the same 5 task orders created by random permutation.
The selected tasks and ordering are listed in the Appendix (Table \ref{tab:dataset_order}).

\paragraph{Hyper-parameters}
We use T5-small as the backbone model and reuse its sentinel tokens \cite{raffel2020t5}.
For each task, \cpt first trains 10 epochs with fused query (and using memory if available) for forward transfer. Afterward, it concentrates on the current task and continues training 10 epochs on the original data of the current task.
When using backward transfer, we train 5 epochs for each previous task.
Other methods train 20 epochs for each task.
We use AdamW and set the learning rate to 3e-5 for \ft, \replay, and \ewc, 3e-3 for \adapter, and 0.5 for all prompt tuning based methods.
We set the batch size to 16 for prompt tuning based methods and 8 for other methods.
To avoid overfitting, we perform early stopping if validation performance does not improve for 5 consecutive epochs.
The weight for \ewc regularization loss is 0.01.
We set the memory size $|M|$ to 50 for each task and save the same samples for all methods that require memory.
We initialize prompt tokens with the tokens randomly drawn from the vocabulary.
For prompt tuning based methods, we tune 100 soft prompt tokens with the embedding size 512 for each task, resulting in 51.2K parameters.
To compare parameter efficiency, we adjust \adapter's parameters for each task to be nearly 1x or 20x as ours.

\section{Experiments and Analysis}
The experiments are organized as follows.
We compare our method with baselines in Sec. \ref{sec:main}, and present a comprehensive ablation study in Sec. \ref{sec:ablation}.
We investigate the effect of prompt initialization in Sec. \ref{sec:exp_init}, and the effect of model size and prompt length in Sec. \ref{sec:model_size}.

\subsection{Main Experiment}
\label{sec:main}

\begin{table*}[t]
\centering
\begin{tabular}{l|ccc|ccc}
\toprule
Method & Avg. JGA & FWT & BWT & Memory & +Params & Tune Params \\
\midrule
\ft             & 14.3$_{0.8}$ &\ \ 8.3$_{1.0}$ & -49.9$_{4.4}$ & - & 0 & 1  \\
\ewc            & 13.9$_{1.1}$ &\ \ 8.4$_{0.9}$ & -50.8$_{4.3}$ & $|M|$*T & 2 & 1 \\
\replay         & 58.6$_{3.5}$ & 10.9$_{0.5}$ &\ \ -3.2$_{2.3}$ & $|M|$*T & 0 & 1 \\
\adapter (20x)  & 49.8$_{1.7}$ & - & - & - & 2\%*T & 2\% \\
\adapter (1x)   & 30.6$_{1.1}$ & - & - & - & 0.1\%*T & 0.1\% \\
\midrule
\pt             & 48.1$_{0.9}$ & - & - & - & \multirow{4}{*}{0.1\%*T} & \multirow{4}{*}{0.1\%}\\
\cpt            & \textbf{59.5$_{1.4}$} &\ \ 9.9$_{0.7}$ & 0& - & & \\
\quad \wmem     & 60.7$_{2.4}$ & \textbf{13.7$_{0.8}$}  & 0& $|M|$*T & & \\
\quad \wmemback & \textbf{61.2$_{2.5}$} & \textbf{13.7$_{0.8}$} &\ \ \ \ \textbf{0.5$_{0.4}$} & $|M|$*T & & \\
\midrule
\multipt        & 64.0$_{1.9}$ & - & - & - & 0.1\% & 0.1\% \\
\bottomrule
\end{tabular}
\caption{Performance and resource usage on 15 tasks CL in 5 random orders. Means and standard variances are reported. "T" is the total number of tasks. "+Param" and "Tune Params" are additional parameters in total and tunable parameters for each task, respectively, measured by the ratio to the pre-trained model's parameters. 
We adjust \adapter's parameters for each task to nearly 1x or 20x parameters of prompt tuning based methods.}
\label{tab:main}
\end{table*}

\paragraph{Computation Resource Analysis.} 
In CL, there is a trade-off between performance and computation resources.
Ideally, we hope to utilize the least amount of computation resources to achieve the best performance.
We take three vital resources into our consideration. \textbf{Memory} saves previous tasks' samples, which may involve privacy issue and requires extra storage.
\textbf{Additional parameters} are the extra parameters we add to our model to cope with different tasks along the CL process, which should be kept to a minimum in order to scale to long task sequences.
\textbf{Tunable parameters} are the trainable parameters when we learn a task, which is important for GPU memory and computation.
We show the usage of these resources in Table \ref{tab:main} (right).
\replay stores $|M|$ samples for each task and does not need extra parameters.
\ewc saves the Fisher information matrix and original parameters, requiring two times additional parameters.
\adapter, \pt, and \cpt require no memory and only add a small number (2\% or 0.1\%) of additional parameters for each task, largely reducing the computational and storage overhead.
Apart from the vanilla form, \cpt can also utilize the memory if available.

\paragraph{CL Performance Analysis.}
Overall CL results of different methods are summarized in Table \ref{tab:main} (left). We have the following findings:
\begin{itemize}[leftmargin=*,itemsep=2pt,topsep=0pt,parsep=0pt]
\item Consistent with \citet{madotto-etal-2021-continual}, both \ft and \ewc suffer from catastrophic forgetting while replaying memory can alleviate the problem to a large extend.
\ft and \ewc have a low Avg. JGA because of the large negative BWT, while \replay improves BWT a lot thus has a high Avg. JGA.
\item Our proposed \pt with masked spans recovering is more parameter efficient than \adapter.
In terms of Avg. JGA, \pt is much better than \adapter with the same size and comparable to \adapter with 20x parameters.
\item Forward transfer through \clinit and query fusion is effective for \pt. 
\cpt improves over \pt significantly and outperforms baselines.
\item When memory is available, our method achieves the best results w.r.t. all metrics, closing the gap between CL and multi-task learning.
Memory improves zero-shot performance (FWT) on new tasks as \replay is better than \ft and \cpt \wmem is better than without memory.
\item Our memory-guided backward transfer effectively utilizes subsequent tasks to help previous tasks.
Although minor, \cpt \wmemback is the only method that exhibits positive BWT.
\end{itemize}

\subsection{Ablation Study}
\label{sec:ablation}
\begin{table}[t]
\normalsize
    \centering
    \setlength{\tabcolsep}{1.35mm}{
    \begin{tabular}{ccccccc}
        \toprule
         & MSR & \clinit & QF & MR & Avg. JGA & FWT \\
        \midrule
        1 &             &               &               &               & 29.6$_{1.2}$ & - \\
        2 &             & \checkmark    &               &               & 41.8$_{2.8}$ &\ \ 6.7$_{0.3}$\\
        3 & \checkmark  &               &               &               & 48.1$_{0.9}$ & - \\
        4 & \checkmark  & \checkmark    &               &               & 57.6$_{2.5}$ &\ \ 9.6$_{1.2}$ \\
        5 & \checkmark  & \checkmark    & \checkmark    &               & 59.5$_{1.4}$ &\ \ 9.9$_{0.7}$ \\
        6 & \checkmark  & \checkmark    &               & \checkmark    & 60.4$_{1.1}$ & 11.9$_{0.6}$ \\
        7 & \checkmark  & \checkmark    & \checkmark    & \checkmark    & 60.7$_{2.4}$ & 13.7$_{0.8}$ \\
        \bottomrule
    \end{tabular}
    }
    \caption{Ablation study for masked spans recovering formulation (MSR), prompt initialization (\clinit or random), query fusion (QF) and memory replay (MR).}
    \label{tab:ablation}
\end{table}

To understand the effect of different proposed techniques, we conduct an in-depth ablation study and show the result in Table \ref{tab:ablation}.
Row 1 and 2 do not formulate DST as a masked spans recovering (MSR) task: the input is the concatenate of the dialog, service name, and soft prompt, while the output is a sequence of slot-value pairs as in \ft (Figure \ref{fig:2}(a)).
Several interesting observations can be noted:
\textbf{First}, formulating DST as MSR is beneficial. Using MSR achieves better CL performance regardless of learning each task independently (row 3 v.s. row 1) or continually using \clinit (row 4 v.s. row 2).
Besides, MSR formulation improves zero-shot generalization on new tasks (row 4 v.s. row 2).
\textbf{Second}, forward transfer through \clinit brings large improvement for CL.
\clinit outperforms random initialization greatly for both using MSR formulation (row 4 v.s. 3) and not (row 2 v.s. 1).
\textbf{Third}, both query fusion and memory replay are effective. 
When they are used separately, memory replay (row 6) boosts the performance more than query fusion (row 5), while applying them altogether achieves the best performance (row 7).

\subsection{Continual Prompt Initialization}
\label{sec:exp_init}

In this experiment (Table \ref{tab:init}), we compare \clinit with other prompt initialization strategies for \pt in CL.
\selectinit (see Sec. \ref{sec:init}) selects the prompt that has the best zero-shot performance on the current task from all previous tasks' prompts for initialization.
We could see that both \selectinit and \clinit outperform random initialization significantly, demonstrating the effectiveness of transferring knowledge from previous tasks through prompt initialization.
\clinit is slightly better than \selectinit in both Avg. JGA and zero-shot generalization (FWT), which reveals the benefit of accumulating knowledge from all seen tasks.
In contrast, the prompt initialized by \selectinit has seen \textit{fewer} tasks and thus contains less knowledge, which might explain the slightly worse result.

\begin{table}[t]
    \centering
    \begin{tabular}{ccc}
    \toprule
        Initialization & Avg. JGA & FWT \\
    \midrule
        Random & 48.1$_{0.9}$ & - \\
        \selectinit & 54.5$_{2.0}$  & 8.2$_{1.3}$ \\
        \clinit & 57.6$_{2.5}$ & 9.6$_{1.2}$ \\
    \bottomrule
    \end{tabular}
    \caption{Comparison of different prompt initialization strategies for \pt.}
    \label{tab:init}
\end{table}

\begin{table}[t]
    \centering
    \begin{tabular}{c|ccc}
        \toprule
        Training & \multicolumn{3}{c}{Testing tasks} \\
        task sequence & $\mathcal{T}_{40:44}$    & $\mathcal{T}_{30:44}$   & $\mathcal{T}_{15:44}$   \\
        \midrule
        $\mathcal{T}_{40:44}$           & 45.1      & -         & -         \\
        $\mathcal{T}_{30:44}$          & 54.2      & 59.7      & -         \\
        $\mathcal{T}_{15:44}$          & 59.0      & 64.4      & 64.3      \\
        $\mathcal{T}_{1:44}$          & 60.7      & 67.8      & 69.3      \\
        \bottomrule
    \end{tabular}
    \caption{\pt with \clinit on the last 5, 15, 30, and 44 (all) tasks of the same task order. We report the Avg. JGA on the last 5, 15, and 30 tasks, respectively.}
    \label{tab:task_num}
\end{table}

Based on the observation above, we further study that whether seeing \textit{more} preceding tasks further helps \clinit.
To this end, we choose a task order of all 44 tasks at random (see Table \ref{tab:dataset_count} in the Appendix) and perform \pt with \clinit on the last 5, last 15, last 30, and all 44 tasks separately.
Formally, we train on four CL curriculums $\mathcal{T}_{40:44}$, $\mathcal{T}_{30:44}$, $\mathcal{T}_{15:44}$, and $\mathcal{T}_{1:44}$, which have the same ending.
We calculate the Avg. JGA on the $\mathcal{T}_{40:44}$, $\mathcal{T}_{30:44}$, and $\mathcal{T}_{15:44}$ if possible.
As illustrated in Table \ref{tab:task_num}, performance on the same tasks (in the same column) increases monotonously as the number of preceding tasks grows. This pattern validates that the benefit of \clinit becomes more evident as the number of tasks increases.
This finding suggests that our method is suitable for long task sequences.

\subsection{Model Size and Prompt Length}
\label{sec:model_size}

In this experiment, we analyze the influence of pre-trained model size and prompt length.
We vary the pre-trained model in \{T5-small, T5-base, T5-large\} and prompt length in \{1, 5, 20, 100, 150\} for \cpt on the 15 tasks (the task order is in Table \ref{tab:dataset_order} in the Appendix).
Figure \ref{fig:model_params} shows Avg. JGA and Table \ref{tab:model_params} shows FWT.
We can observe that:
\textbf{First}, when fixing the prompt length, increasing the model size improves the Avg. JGA as well as the generalization ability measured by FWT in most cases.
\textbf{Second}, when the backbone model size is fixed, increasing the prompt length improves the overall performance in general. 
Furthermore, we found that increasing prompt token length from 20 to 100 improves Avg. JGA and FWT more than increasing it from 100 to 150, which is consistent with the finding in \citet{lester-etal-2021-power}.
\textbf{Third}, our method becomes more parameter-efficient as the backbone model size grows. With the same number of tunable parameters (x-axis), using a larger pre-trained model achieves better Avg. JGA.

\begin{figure}[t]
    \centering
    \includegraphics[width=\linewidth]{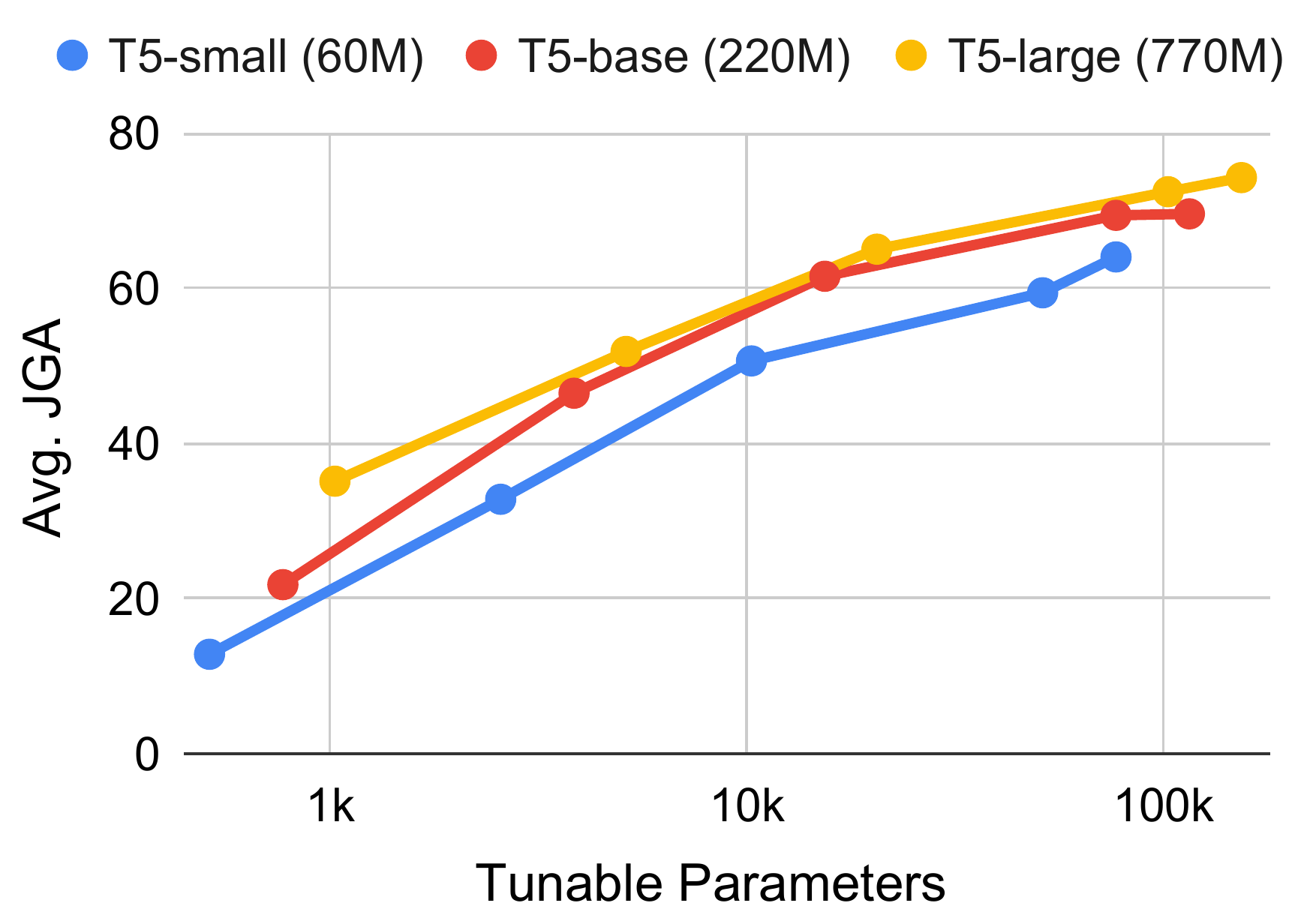}
    \caption{Avg. JGA for \cpt with different pre-trained models and prompt lengths. The x-axis is the number of tunable parameters in log scale. The points on each curve correspond to 1, 5, 20, 100, and 150 prompt tokens from left to right.}
    \label{fig:model_params}
\end{figure}

\begin{table}[t]
    \centering
    \setlength{\tabcolsep}{1.5mm}{
    \begin{tabular}{lrrrrr}
    \toprule
                                & \multicolumn{5}{c}{Prompt Length} \\
    \cmidrule(lr){2-6}
                            & 1   & 5    & 20     & 100    & 150 \\
    \midrule
        T5-small (60M)   & \ \ 6.1 & \ \ 6.7 & \ \ 8.9 &\ \ 9.8 &\ \ 9.8\\
        T5-base (220M)   & \ \ 5.7 & \ \ 9.9 & 12.9   & 18.3   & 15.0\\
        T5-large (770M) & 10.6 & 17.0 & 18.5   & 28.0   & 31.2\\
    \bottomrule
    \end{tabular}
    }
    \caption{FWT for \cpt with different pre-trained models and prompt lengths.}
    \label{tab:model_params}
\end{table}

\subsection{The Effect of Memory Size}
\label{sec:mem}

In this section, we compare the role of memory in \replay and our method.
We vary the memory size per task $|M|$ in \{10, 50, 100\} and show the performance of \replay and \cpt with memory replay (and memory-guided backward transfer) in Table \ref{tab:mem_size}.
We can find that increasing the memory size benefits \replay significantly.
This is not surprising because \replay and other rehearsal methods rely on memory to solve the challenging forgetting problem.
When the memory size is unlimited, \replay degenerates to multi-task learning, which is powerful but costly in storage and computation.

\begin{table}[t]
    \centering
    \setlength{\tabcolsep}{1.3mm}{
    \begin{tabular}{lcccc}
    \toprule
                            & \multicolumn{3}{c}{Memory Size} \\
    \cmidrule(lr){2-4}
                            & 10           & 50           & 100   \\
    \midrule
        \replay             & 44.0$_{1.0}$ & 58.6$_{3.5}$ & 65.6$_{0.8}$\\
        \textit{CPT w/ mem.}& 59.0$_{3.3}$ & 60.7$_{2.4}$ & 59.7$_{3.2}$\\
        \textit{CPT w/ mem. \& back.}& 58.6$_{3.7}$ & 61.2$_{2.5}$ & 60.4$_{3.3}$\\
        \quad BWT & \ -0.4$_{0.5}$ & \ \ 0.5$_{0.4}$ & \ \ 0.8$_{0.4}$ \\
    \bottomrule
    \end{tabular}
    }
    \caption{Avg. JGA for \replay and \cpt (\textit{CPT}) with memory replay (and memory-guided backward transfer) using different memory size. BWT for \textit{CPT w/ mem. \& back.} is also shown.}
    \label{tab:mem_size}
    \vspace{-1em}
\end{table}

For \cpt, however, the memory is not used for retaining the performance on previous tasks since parameters for previous tasks are saved.
\begin{itemize}[leftmargin=*,itemsep=2pt,topsep=0pt,parsep=0pt]
\item In forward transfer, the memory helps recall previous tasks' knowledge and serves as a complement to \clinit and query fusion.
The influence on Avg. JGA depends on the effect of transfer learning on the current task via multi-task training ($\mathcal{L}_{\theta_{P_k}}(D_k+M_{<k})$).
As shown in the row 2 in Table \ref{tab:mem_size}, increasing the memory size does not improve Avg. JGA significantly and may even distract the model from learning the current domain.
This result suggests that our method does not need a large memory for forward transfer.
\item In backward transfer, the memory gives reference gradients to guide the updates and serves as a filter to decide whether to accept the updates.
Thus larger memory gives more accurate guidance.
From the bottom row in Table \ref{tab:mem_size}, we can find that increasing memory size can improve the effect of backward transfer.
\end{itemize}

We also conduct experiments using a percentage memory budget, setting the memory size for each task proportional to task data size: $|M_i| \propto |D_i|$.
This means low-resource tasks have fewer samples stored in the memory than in the original setting.
We set the total memory size to 50 * T, where T is the number of tasks.
As shown in Table \ref{tab:proportional_mem_size}, \replay performs much worse (58.6$\rightarrow$55.8) in the unbalanced task memory setting while the effect on \cpt \textit{w/ mem.} is slight (60.7$\rightarrow$60.3).
Besides, our proposed backward transfer technique is still effective.

\begin{table}[t]
    \centering
    \setlength{\tabcolsep}{1.3mm}{
    \begin{tabular}{lccc}
    \toprule
                            & \multicolumn{2}{c}{Memory Size} \\
    \cmidrule(lr){2-3}
                            & fixed = 50           & proportional           \\
    \midrule
        \replay             & 58.6$_{3.5}$ & 55.8$_{0.7}$\\
        \textit{CPT w/ mem.}& 60.7$_{2.4}$ & 60.3$_{3.1}$\\
        \textit{CPT w/ mem. \& back.}& 61.2$_{2.5}$ & 60.7$_{3.4}$\\
        \quad BWT &  \ \ 0.5$_{0.4}$ & \ \ 0.4$_{0.5}$ \\
    \bottomrule
    \end{tabular}
    }
    \caption{Avg. JGA for \replay and \cpt (\textit{CPT}) with memory replay (and memory-guided backward transfer) using the fixed/proportional memory size. The total memory sizes are the same. BWT for \textit{CPT w/ mem. \& back.} is also shown.}
    \label{tab:proportional_mem_size}
    \vspace{-1em}
\end{table}

Overall, these results indicate that compared with \replay, our method uses the memory differently and benefits less from enlarging the memory.

\section{Conclusion}
In this paper, we develop prompt tuning for continual learning for the first time.
We propose \cpt, a highly parameter-efficient framework that avoids forgetting and enables forward/backward knowledge transfer among tasks. 
For forward transfer, we explore continual prompt initialization, query fusion, and memory replay techniques.
For backward transfer, we devise a memory-guided technique.
Extensive experiments on continual learning for DST demonstrate the effectiveness and efficiency of our proposed method compared with state-of-the-art baselines.
Our method and findings will foster more future studies towards building more scalable, adaptable task-oriented dialog systems.

\section*{Acknowledgements}
This work was supported by the National Science Foundation for Distinguished Young Scholars (with No. 62125604) and the NSFC projects (Key project with No. 61936010 and regular project with No. 61876096). This work was also supported by the Guoqiang Institute of Tsinghua University, with Grant No. 2019GQG1 and 2020GQG0005.

\bibliography{anthology,custom}
\bibliographystyle{acl_natbib}

\begin{table*}[t]
\centering
\begin{tabular}{ccrrrrrrrrr}
\toprule
Task ID &
  Service &
  \multicolumn{1}{c}{\# Slots} &
  \multicolumn{3}{c}{\# Dialogs} &
  \multicolumn{3}{c}{\# Samples} &
  \multicolumn{2}{c}{Avg. tokens} \\ \midrule
\textit{} &
  \textit{} &
  \multicolumn{1}{c|}{\textit{}} &
  \multicolumn{1}{c}{\textit{Train}} &
  \multicolumn{1}{c}{\textit{Dev}} &
  \multicolumn{1}{c|}{\textit{Test}} &
  \multicolumn{1}{c}{\textit{Train}} &
  \multicolumn{1}{c}{\textit{Dev}} &
  \multicolumn{1}{c|}{\textit{Test}} &
  \multicolumn{1}{c}{\textit{Context}} &
  \multicolumn{1}{c}{\textit{Query}} \\ \midrule
1  & events\_3      & \multicolumn{1}{c|}{5}  & 53  & 7  & \multicolumn{1}{r|}{16}  & 312  & 40  & \multicolumn{1}{r|}{105}  & 121 & 47 \\
2  & banks\_2       & \multicolumn{1}{c|}{4}  & 29  & 4  & \multicolumn{1}{r|}{9}   & 220  & 31  & \multicolumn{1}{r|}{72}   & 111 & 49 \\
3  & banks\_1       & \multicolumn{1}{c|}{4}  & 144 & 21 & \multicolumn{1}{r|}{42}  & 1138 & 169 & \multicolumn{1}{r|}{335}  & 114 & 57 \\
4  & calendar\_1    & \multicolumn{1}{c|}{4}  & 118 & 17 & \multicolumn{1}{r|}{34}  & 773  & 110 & \multicolumn{1}{r|}{234}  & 112 & 33 \\
5  & movies\_3      & \multicolumn{1}{c|}{3}  & 33  & 5  & \multicolumn{1}{r|}{10}  & 112  & 18  & \multicolumn{1}{r|}{37}   & 72  & 26 \\
6  & music\_2       & \multicolumn{1}{c|}{5}  & 231 & 33 & \multicolumn{1}{r|}{67}  & 1593 & 221 & \multicolumn{1}{r|}{469}  & 117 & 54 \\
7  & services\_2    & \multicolumn{1}{c|}{5}  & 129 & 19 & \multicolumn{1}{r|}{37}  & 917  & 148 & \multicolumn{1}{r|}{253}  & 131 & 52 \\
8  & payment\_1     & \multicolumn{1}{c|}{4}  & 25  & 3  & \multicolumn{1}{r|}{8}   & 233  & 33  & \multicolumn{1}{r|}{89}   & 171 & 52 \\
9  & media\_1       & \multicolumn{1}{c|}{4}  & 196 & 28 & \multicolumn{1}{r|}{57}  & 1207 & 182 & \multicolumn{1}{r|}{360}  & 99  & 48 \\
10 & weather\_1     & \multicolumn{1}{c|}{2}  & 58  & 8  & \multicolumn{1}{r|}{17}  & 259  & 39  & \multicolumn{1}{r|}{66}   & 77  & 16 \\
11 & events\_1      & \multicolumn{1}{c|}{6}  & 202 & 29 & \multicolumn{1}{r|}{58}  & 1424 & 195 & \multicolumn{1}{r|}{400}  & 132 & 64 \\
12 & flights\_4     & \multicolumn{1}{c|}{7}  & 60  & 9  & \multicolumn{1}{r|}{18}  & 290  & 41  & \multicolumn{1}{r|}{87}   & 90  & 77 \\
13 & travel\_1      & \multicolumn{1}{c|}{4}  & 48  & 7  & \multicolumn{1}{r|}{14}  & 231  & 28  & \multicolumn{1}{r|}{63}   & 87  & 59 \\
14 & buses\_2       & \multicolumn{1}{c|}{6}  & 111 & 16 & \multicolumn{1}{r|}{32}  & 857  & 120 & \multicolumn{1}{r|}{234}  & 137 & 54 \\
15 & events\_2      & \multicolumn{1}{c|}{6}  & 400 & 57 & \multicolumn{1}{r|}{115} & 3537 & 521 & \multicolumn{1}{r|}{1067} & 159 & 59 \\
16 & alarm\_1       & \multicolumn{1}{c|}{2}  & 58  & 9  & \multicolumn{1}{r|}{17}  & 367  & 49  & \multicolumn{1}{r|}{107}  & 101 & 22 \\
17 & buses\_3       & \multicolumn{1}{c|}{7}  & 61  & 9  & \multicolumn{1}{r|}{18}  & 405  & 66  & \multicolumn{1}{r|}{114}  & 123 & 69 \\
18 & services\_1    & \multicolumn{1}{c|}{5}  & 185 & 27 & \multicolumn{1}{r|}{53}  & 1241 & 180 & \multicolumn{1}{r|}{352}  & 129 & 58 \\
19 & buses\_1       & \multicolumn{1}{c|}{5}  & 136 & 20 & \multicolumn{1}{r|}{39}  & 1054 & 143 & \multicolumn{1}{r|}{313}  & 138 & 49 \\
20 & restaurants\_2 & \multicolumn{1}{c|}{9}  & 87  & 13 & \multicolumn{1}{r|}{28}  & 807  & 113 & \multicolumn{1}{r|}{240}  & 154 & 97 \\
21 & hotels\_2      & \multicolumn{1}{c|}{6}  & 212 & 31 & \multicolumn{1}{r|}{61}  & 1569 & 234 & \multicolumn{1}{r|}{460}  & 152 & 73 \\
22 & ridesharing\_2 & \multicolumn{1}{c|}{3}  & 64  & 9  & \multicolumn{1}{r|}{19}  & 380  & 49  & \multicolumn{1}{r|}{108}  & 106 & 34 \\
23 & rentalcars\_1  & \multicolumn{1}{c|}{6}  & 100 & 14 & \multicolumn{1}{r|}{29}  & 840  & 120 & \multicolumn{1}{r|}{242}  & 161 & 59 \\
24 & movies\_1      & \multicolumn{1}{c|}{8}  & 263 & 37 & \multicolumn{1}{r|}{76}  & 1873 & 250 & \multicolumn{1}{r|}{556}  & 122 & 70 \\
25 & ridesharing\_1 & \multicolumn{1}{c|}{3}  & 74  & 10 & \multicolumn{1}{r|}{22}  & 412  & 57  & \multicolumn{1}{r|}{125}  & 103 & 36 \\
26 & media\_3       & \multicolumn{1}{c|}{4}  & 56  & 8  & \multicolumn{1}{r|}{16}  & 327  & 42  & \multicolumn{1}{r|}{89}   & 95  & 36 \\
27 & music\_3       & \multicolumn{1}{c|}{6}  & 17  & 3  & \multicolumn{1}{r|}{5}   & 112  & 19  & \multicolumn{1}{r|}{32}   & 114 & 60 \\
28 & movies\_2      & \multicolumn{1}{c|}{3}  & 32  & 5  & \multicolumn{1}{r|}{10}  & 118  & 20  & \multicolumn{1}{r|}{38}   & 70  & 30 \\
29 & flights\_2     & \multicolumn{1}{c|}{7}  & 129 & 19 & \multicolumn{1}{r|}{37}  & 822  & 115 & \multicolumn{1}{r|}{251}  & 127 & 75 \\
30 & services\_4    & \multicolumn{1}{c|}{5}  & 86  & 13 & \multicolumn{1}{r|}{25}  & 680  & 97  & \multicolumn{1}{r|}{208}  & 154 & 49 \\
31 & flights\_1     & \multicolumn{1}{c|}{10} & 560 & 80 & \multicolumn{1}{r|}{160} & 4680 & 667 & \multicolumn{1}{r|}{1379} & 168 & 10 \\
32 & services\_3    & \multicolumn{1}{c|}{5}  & 131 & 19 & \multicolumn{1}{r|}{38}  & 959  & 143 & \multicolumn{1}{r|}{290}  & 143 & 54 \\
33 & flights\_3     & \multicolumn{1}{c|}{8}  & 65  & 10 & \multicolumn{1}{r|}{19}  & 420  & 75  & \multicolumn{1}{r|}{116}  & 133 & 79 \\
34 & trains\_1      & \multicolumn{1}{c|}{7}  & 58  & 9  & \multicolumn{1}{r|}{17}  & 415  & 67  & \multicolumn{1}{r|}{117}  & 131 & 76 \\
35 & homes\_2       & \multicolumn{1}{c|}{8}  & 62  & 9  & \multicolumn{1}{r|}{18}  & 424  & 56  & \multicolumn{1}{r|}{139}  & 140 & 89 \\
36 & rentalcars\_2  & \multicolumn{1}{c|}{6}  & 77  & 11 & \multicolumn{1}{r|}{23}  & 631  & 91  & \multicolumn{1}{r|}{185}  & 157 & 61 \\
37 & restaurants\_1 & \multicolumn{1}{c|}{9}  & 256 & 37 & \multicolumn{1}{r|}{74}  & 2098 & 297 & \multicolumn{1}{r|}{581}  & 153 & 10 \\
38 & music\_1       & \multicolumn{1}{c|}{6}  & 68  & 10 & \multicolumn{1}{r|}{20}  & 468  & 73  & \multicolumn{1}{r|}{142}  & 118 & 61 \\
39 & hotels\_4      & \multicolumn{1}{c|}{7}  & 80  & 12 & \multicolumn{1}{r|}{23}  & 559  & 99  & \multicolumn{1}{r|}{141}  & 134 & 72 \\
40 & media\_2       & \multicolumn{1}{c|}{5}  & 32  & 4  & \multicolumn{1}{r|}{10}  & 215  & 29  & \multicolumn{1}{r|}{71}   & 112 & 59 \\
41 & hotels\_3      & \multicolumn{1}{c|}{6}  & 90  & 13 & \multicolumn{1}{r|}{26}  & 737  & 100 & \multicolumn{1}{r|}{193}  & 157 & 64 \\
42 & rentalcars\_3  & \multicolumn{1}{c|}{7}  & 44  & 7  & \multicolumn{1}{r|}{13}  & 332  & 55  & \multicolumn{1}{r|}{99}   & 148 & 72 \\
43 & hotels\_1      & \multicolumn{1}{c|}{7}  & 99  & 14 & \multicolumn{1}{r|}{29}  & 868  & 105 & \multicolumn{1}{r|}{250}  & 161 & 71 \\
44 & homes\_1       & \multicolumn{1}{c|}{7}  & 244 & 35 & \multicolumn{1}{r|}{70}  & 1829 & 282 & \multicolumn{1}{r|}{540}  & 159 & 81  \\ 
\bottomrule
\end{tabular}%
\caption{Statistics of the services we used. 
Average tokens of dialog context and query is calculated using T5 tokenizer.
Services are arranged in the order of their appearance in our 44 task experiment (Sec. \ref{sec:exp_init}).
Last 15 services are used for all our 15 task experiments.
}
\label{tab:dataset_count}
\end{table*}

\begin{table*}[t]
\centering
\begin{tabular}{c|ccccccccccccccc}
\toprule
Task order & \multicolumn{15}{c}{Tasks' IDs in order} \\
\midrule
Order1 & 30 & 31 & 32 & 33 & 34 & 35 & 36 & 37 & 38 & 39 & 40 & 41 & 42 & 43 & 44 \\
Order2 & 39 & 33 & 36 & 42 & 40 & 37 & 38 & 34 & 32 & 35 & 41 & 31 & 30 & 44 & 43 \\
Order3 & 30 & 41 & 38 & 31 & 43 & 39 & 40 & 33 & 34 & 44 & 37 & 36 & 32 & 35 & 42 \\
Order4 & 43 & 40 & 44 & 38 & 30 & 37 & 31 & 39 & 32 & 35 & 41 & 34 & 33 & 36 & 42 \\
Order5 & 30 & 33 & 44 & 31 & 38 & 32 & 42 & 40 & 37 & 43 & 36 & 39 & 41 & 35 & 34 \\ \bottomrule
\end{tabular}%
\caption{Five task orders of all our 15 tasks experiments.
We use last 15 tasks in Table \ref{tab:dataset_count}. 
The task order for Section \ref{sec:model_size} is \textit{Order1}.
}
\label{tab:dataset_order}
\end{table*}

\end{document}